\documentclass[a4paper,fleqn]{cas-dc}

\usepackage[numbers]{natbib}
\usepackage{times}
\usepackage{soul}
\usepackage{url}
\usepackage{hyperref}
\usepackage{caption}
\usepackage{graphicx}
\usepackage{amsmath}
\usepackage{amsthm}
\usepackage{booktabs}
\usepackage{algorithm}
\usepackage{algorithmic}
\usepackage[switch]{lineno}
\usepackage{subfig}
\usepackage{textcomp}
\usepackage{amsfonts}
\usepackage{color}
\usepackage{multirow}
\usepackage{float}

\def\tsc#1{\csdef{#1}{\textsc{\lowercase{#1}}\xspace}}
\tsc{WGM}
\tsc{QE}

\begin{document}
\let\WriteBookmarks\relax
\def\floatpagepagefraction{1}
\def\textpagefraction{.001}
\let\printorcid\relax

\shorttitle{DHRNet: A Dual-Path Hierarchical Relation Network for Multi-Person Pose Estimation}

\shortauthors{Yonghao Dang et al.}

\title [mode = title]{DHRNet: A Dual-Path Hierarchical Relation Network for Multi-Person Pose Estimation}  

%

\author[1]{Yonghao Dang}
\ead{dyh2018@bupt.edu.cn}

\author[1]{Jianqin Yin}
\cormark[1]
\ead{jqyin@bupt.edu.cn}

\author[1]{Liyuan Liu}

\author[2]{Pengxiang Ding}

\author[3]{Yuan Sun}

\author[4]{Yanzhu Hu}

\affiliation[1]{organization={School of Artificial Intelligence, Beijing University of Posts and Telecommunications},
            city={Beijing},
            postcode={100876}, 
            country={China}}

\affiliation[2]{organization={School of Engineering, Westlake University},
            city={Hangzhou},
            postcode={310024}, 
            country={China}}
            
\affiliation[3]{organization={School of Electronic Engineering, Beijing University of Posts and Telecommunications},
            city={Beijing},
            postcode={100876}, 
            country={China}}

\affiliation[4]{organization={School of Modern Post (School of Automation), Beijing University of Posts and Telecommunications},
            city={Beijing},
            postcode={100876}, 
            country={China}}

\cortext[1]{Corresponding author}



\begin{abstract}
 Multi-person pose estimation (MPPE) presents a formidable yet crucial challenge in computer vision. Most existing methods predominantly concentrate on isolated interaction either between instances or joints, which is inadequate for scenarios demanding concurrent localization of both instances and joints. This paper introduces a novel CNN-based single-stage method, named Dual-path Hierarchical Relation Network (DHRNet), to extract instance-to-joint and joint-to-instance interactions concurrently. Specifically, we design a dual-path interaction modeling module (DIM) that strategically organizes cross-instance and cross-joint interaction modeling modules in two complementary orders, enriching interaction information by integrating merits from different correlation modeling branches. Notably, DHRNet excels in joint localization by leveraging information from other instances and joints. Extensive evaluations on challenging datasets, including COCO, CrowdPose, and OCHuman datasets, showcase DHRNet's state-of-the-art performance. The code will be released at \href{https://github.com/YHDang/dhrnet-multi-pose-estimation}{https://github.com/YHDang/dhrnet-multi-pose-estimation}.
\end{abstract}






\maketitle

\section{Introduction}
 Multi-person pose estimation (MPPE) stands as a pivotal task in computer vision, focusing on detecting keypoints for each individual within an image. Its applications span diverse fields, including teaching management and evaluation \cite{gao2021multi}, human-robot interaction \cite{HRInteraction}, and virtual reality \cite{Tsinghua}. Recent strides in deep learning have markedly advanced MPPE.

\begin{figure}
    \centering
    \includegraphics[scale=0.4]{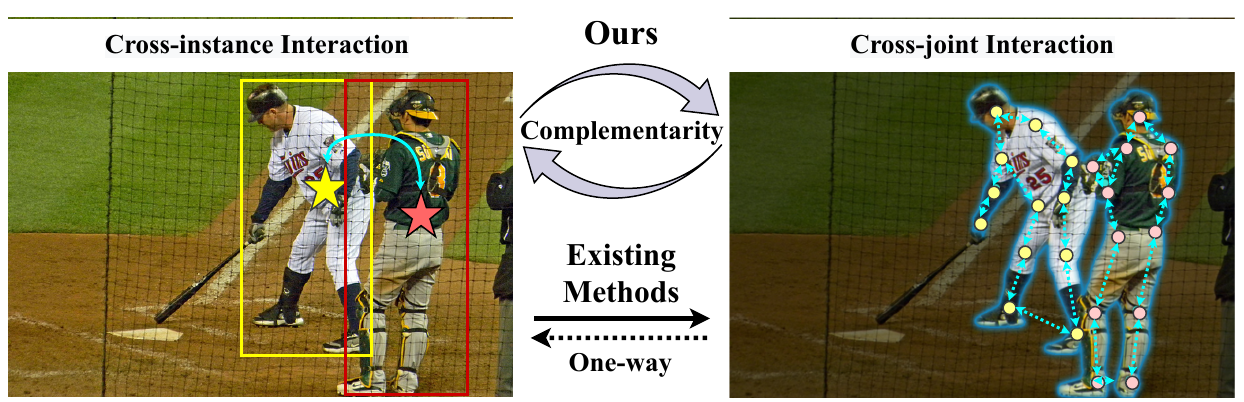}
    \caption{Interactive information in multi-person scenario. The solid double arrows represent cross-instance interaction. The dashed double arrows denote cross-joint interaction. Most existing methods model the interactive information with a one-way order, which ignores the complementarity between cross-instance and cross-joint interactions. Our method makes full use of the complementarity of these two interactions through bidirectional correlation modeling.}
    \label{fig:fig1_motivation}
\end{figure}

 While prevalent multi-person estimation methods have achieved promising performance with both top-down \cite{SBL,HRNet,OPEC-Net,TFPose,PPT,SwinPose,RPSTN} and bottom-up \cite{HGG,HigherHRNet,SWAHR} methods, they often face challenges in concurrently capturing cross-instance (\emph{i.e.}, instance-to-instance) and cross-joint (\emph{i.e.}, joint-to-joint) interactions due to their non-differentiable paradigms. In multi-person scenarios, rich interactive information, encompassing instance-to-instance and joint-to-joint interactions (as shown in Figure \ref{fig:fig1_motivation}), plays a crucial role. Cross-instance interaction aids in locating the current person with information from others, while cross-joint interaction helps pinpoint the current joint with insights from other joints. Thus, these two interactive pieces of information are complementary to each other. Leveraging both interactive information sources becomes essential to enhance the performance of multi-person pose estimation methods.
 
 Some single-stage methods have recently been proposed to capture interactive relationships within multi-person scenarios \cite{PRTR,PETR,EDPose,GroupPose}. Unlike two-stage methods, these single-stage approaches unify object and keypoint detection in a compact framework, enabling flexible feature extraction in an end-to-end manner. While these methods have achieved commendable results, they typically adhere to a single fixed order for modeling interactive information, either through instance-to-joint \cite{PRTR,PETR,EDPose} or joint-to-instance \cite{GroupPose,I2RNet} correlations, as illustrated in Figure \ref{fig:fig1_motivation}. This singular approach tends to overlook the complementarity between cross-instance and cross-joint interactions. It is often difficult to model complex interactions from a single view of instance-to-joint or joint-to-instance correlation modeling.

 In this paper, we propose a pioneering single-stage method called \textbf{D}ual-path \textbf{H}ierarchical \textbf{R}elation \textbf{Net}work (DHRNet) designed to explore interactive information between instances or joints simultaneously (refer to Figure \ref{fig:fig2_overview}). Our approach leverages the complementary nature of cross-instance and cross-joint interactions to enhance multi-person pose estimation performance. To achieve this object, we employ an instance decoder and a keypoints decoder to distill representations of instances and joints from the visual features extracted by the backbone. Based on these decoupled features, we introduce a dual-path interaction modeling module (DIM) explicitly designed to model cross-instance and cross-joint interactions. DIM comprises two branches, organizing cross-instance (CIM) and cross-joint (CJM) interaction modeling modules in two complementary orders to extract instance-to-joint and joint-to-instance interaction information. Cross-instance interaction pays attention to the person who is important for detecting the current human. In contrast, cross-joint interaction focuses on the joint that is significant for locating the current one. Additionally, DIM incorporates two adaptive feature fusion modules (ADFMs) to enhance the communication between these two forms of interactive information. To fully leverage these complementary sources, we introduce an adaptive pose decoder to fuse and activate essential features for locating human joints using spatial and channel-wise attention. 
 
 The main contributions of this paper can be summarized as follows.
\begin{itemize}
    \item We investigate a dual-path interaction modeling module (DIM) to make full use of both the cross-instance and cross-joint interactive information through a complementary relation modeling order, enabling the model to locate invisible joints by integrating information from other instances and joints. 
    \item We propose a novel CNN-based Dual-path Hierarchical Relation Network (DHRNet), leveraging cross-instance and cross-joint interactions with an end-to-end framework to handle complex scenarios effectively.
    \item The proposed method achieves new state-of-the-art results on the challenging benchmarks, including COCO\cite{COCO}, CrowdPose\cite{CrowdPose}, and OCHuman\cite{OCHuman} datasets, which demonstrates the effectiveness of our approach.
\end{itemize}

\section{Related Works}

\subsection{Multi-person Pose Estimation}
 \textbf{Top-down methods} typically incorporate an object detector for person detection and a single-person pose estimator to pinpoint joints for each person. Classical pose estimation frameworks such as \cite{SBL,HRNet,CFN,skeletonpose,Smart-VPose} have demonstrated exceptional performance. Fang et al.\cite{RMPE} introduced a regional multi-person pose estimation (RMPE) network to enhance joint localization in the case of erroneous person detection. Qiu et al.\cite{OPEC-Net} designed OPEC-Net, a novel solution facing occlusion challenges by capturing image context and pose structure information. While these methods yield impressive results, they often overlook interactions between different instances.

 \textbf{Bottom-up methods} initially detect all joints and then associate them with the corresponding human. Part affinity fields \cite{Openpose,PifPaf,PersonLab} are commonly used to encode the connection between adjacent joints. Cheng et al.\cite{HigherHRNet} proposed HigherHRNet, utilizing associative embedding for joint grouping. Luo et al.\cite{SWAHR} proposed a novel scale-adaptive heatmap regression method to address multi-scale challenges. Geng et al.\cite{DEKR} proposed DEKR, leveraging multi-branch adaptive convolutions for feature extraction in joint grouping.

 \textbf{Single-stage methods} focus on directly regressing joint positions through an end-to-end approach. Nie et al.\cite{SPM} introduced SPM, which directly regresses joint coordinates. To enhance task-relevant features, Wei et al.\cite{PointPose} proposed a compact point set representation for addressing challenges in pose estimation. Mao et al.\cite{FCPose} employed dynamic keypoint heads to determine joint positions conditioned on each instance representation. The robustness of instance representations is crucial for accurate joint localization. Shi et al.\cite{InsPose} presented instance-aware keypoint networks to directly estimate body pose for each instance. Wang et al.\cite{CID} proposed CID to decouple the instance representations and joint representations. While these methods implicitly capture interactive information between different targets by expanding the receptive field, they may lead to vagueness in the extracted interactive information.

\subsection{Relation Modeling in Pose Estimation}

 Relational information is crucial in accurately locating joints, as the correlations between joints contribute to learning pose structural information \cite{RPSTN}. \cite{TFPose,SwinPose,PPT} used a Transformer to model joint interaction by modeling the correlation between joint tokens. Li et al.\cite{PRTR} proposed PRTR, employing a cascade Transformer for human detection and joint regression. Shi et al. \cite{PETR} framed pose estimation as a hierarchical set prediction problem and presented PETR to regress instance-aware poses directly. Yang et al. \cite{EDPose} proposed ED-Pose, predicting human boxes and joints by first modeling cross-instance correlations followed by cross-joint correlations. Although these recent works use Transformer to calculate relations, their performance is constrained by the lack of correlational information exchange. In contrast, our work simultaneously models coarse-to-fine and fine-to-coarse interactive information with two interaction modeling bran 
 ches, and enrich interaction information by integrating merits from different branches.

\section{Methodology}
\begin{figure*}[htbp]
    \centering
    \includegraphics[scale=0.29]{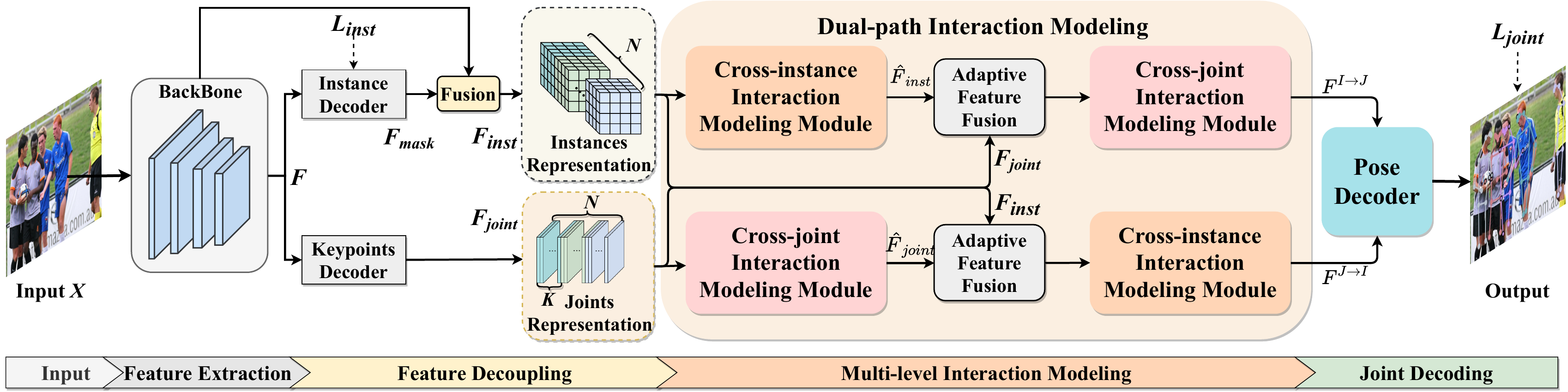}
    \caption{Overview of the proposed DHRNet. For a given input image, a feature encoder is used to extract visual features $F$. Then, an instance decoder and a keypoints decoder are used to generate instance masks $F_{mask}$ and joint representations $F_{joint}$. Instance masks are fused with visual to get instance representations $F_{inst}$. DIM takes $F_{inst}$ and $F_{joint}$ as input to model the correlations. Finally, a pose decoder aggregates relation-based features extracted by DIM to estimate poses.}
    \label{fig:fig2_overview}
\end{figure*}

\subsection{Preliminaries}
 Given an image $X \in \mathbb{R}^{3 \times H \times W}$, where $H$ and $W$ represent the height and width of the image,  containing $N$ instances (\emph{i.e.}, the person), the proposed model aims to detect joints of the $n$-th instance. The $n$-th person's joints can be represented by joint heatmaps $\mathcal{H}^n = \{\mathbf{H}^n_1, \mathbf{H}^n_2, \cdots,\mathbf{H}^n_K \}$, where $\mathbf{H}^n_k \in \mathbb{R}^{h \times w}$ denotes the $k$-th joint heatmap with the size of $h \times w$, and $K$ is the number of joints.

\subsection{Dual-path Hierarchical Relation Network}
 As shown in Figure \ref{fig:fig2_overview}, the framework takes an image as input and employs a pre-trained feature encoder to extract visual features of the image, denoted as $F \in \mathbb{R}^{c \times h \times w}$ (where $c$ denotes the number of channels). The visual feature $F$ amalgamates information about the background and the human. The presence of task-irrelevant background poses a challenge for accurate pose estimation. To effectively model the cross-instance and cross-joint interactions, an instance decoder and a joint decoder are utilized to distill instance-aware representations $F_{inst}=\{F_{inst}^1, F_{inst}^2, \cdots, F_{inst}^N \}$ (where $F_{inst}^n \in \mathbb{R}^{d \times h \times w}$ represents the $n$-th person's feature with dimension $d$) and joint-aware representations $F_{joint}=\{F_{joint}^1, F_{joint}^2, \cdots, F_{joint}^N \}$ (where $F_{joint}^n \in \mathbb{R}^{K \times h \times w}$ denotes $K$ joint features of person $n$).

 $F_{inst}$ and $F_{joint}$ contain information about individual humans and independent joints. A dual-path interaction modeling module (DIM) takes $F_{inst}$ and $F_{joint}$ as input to model cross-instance and cross-joint interactions. As shown in Figure \ref{fig:fig2_overview}, DIM comprises two branches where cross-instance interaction modeling modules (CIM) and cross-joint interaction modeling modules (CJM) are arranged with complementary orders. Since the operations in the two branches in DIM are similar, we illustrate the details using the instance-joint relational branch (IJR) as an example.

 CIM takes the instance representation $F_{inst}$ as input to model interactions between instances (details are in section \ref{sec:DHRM}). Since $F_{inst}$ lacks the information about joints, the joint representation $F_{joint}$ is fused with cross-instance interaction via a channel-wise attention-based adaptive feature fusion module (ADFM) to enhance joint information. Subsequently, the CJM is employed to model cross-joint interactions. The interactive features learned by the IJR branch ($F^{I \rightarrow J}$) can be calculated as follows.
\begin{equation}
    \begin{array}{l}
    \hat{F}_{inst} = CIM(F_{inst}) \\
    F^{I \rightarrow J} = CJM(\mathcal{CA}_R(Conv([\hat{F}_{inst}, F_{joint}])))
    \end{array}
    \label{eq:IJR}
\end{equation}
 where $\hat{F}_{inst}$ denotes cross-instance correlation. Channel-wise attention block $\mathcal{CA}_R(\cdot)$ and a convolutional layer $Conv(\cdot)$ comprise the adaptive feature fusion module. $\left[\cdot,\cdot\right]$ denotes the concatenation operation. Since $F^{I \rightarrow J}$ is a feature obtained firstly by CIM and then CJM, it contains instance-level and joint-level interactive information. Thus, $F^{I \rightarrow J}$ can be regarded as coarse-to-fine interactive information. The principle of the joint-instance relational branch (JIR) is similar to that of IJR. The features $F^{J \rightarrow I}$ extracted by JIR can be viewed as fine-to-coarse interactive information.

 To fully leverage complementary interactive information, the features from both IJR and JIR branches are fed into a pose decoder $\mathcal{D}_p$ to generate robust pose representations.
\begin{equation}
    \mathcal{H}=\mathcal{D}_p([F^{I \rightarrow J}, F^{J \rightarrow I}])
    \label{eq:posedecoder}
\end{equation}

\subsection{Dual-path Interaction Modeling Module}
\label{sec:DHRM}
 The dual-path interaction modeling module (DIM) contains three components: the cross-instance interaction modeling module (CIM), the cross-joint interaction modeling module (CJM), and the adaptive fusion module (ADFM).

\subsubsection{Cross-instance Interaction Modeling Module}

\begin{figure}[htbp]
    \centering
    \includegraphics[scale=0.3]{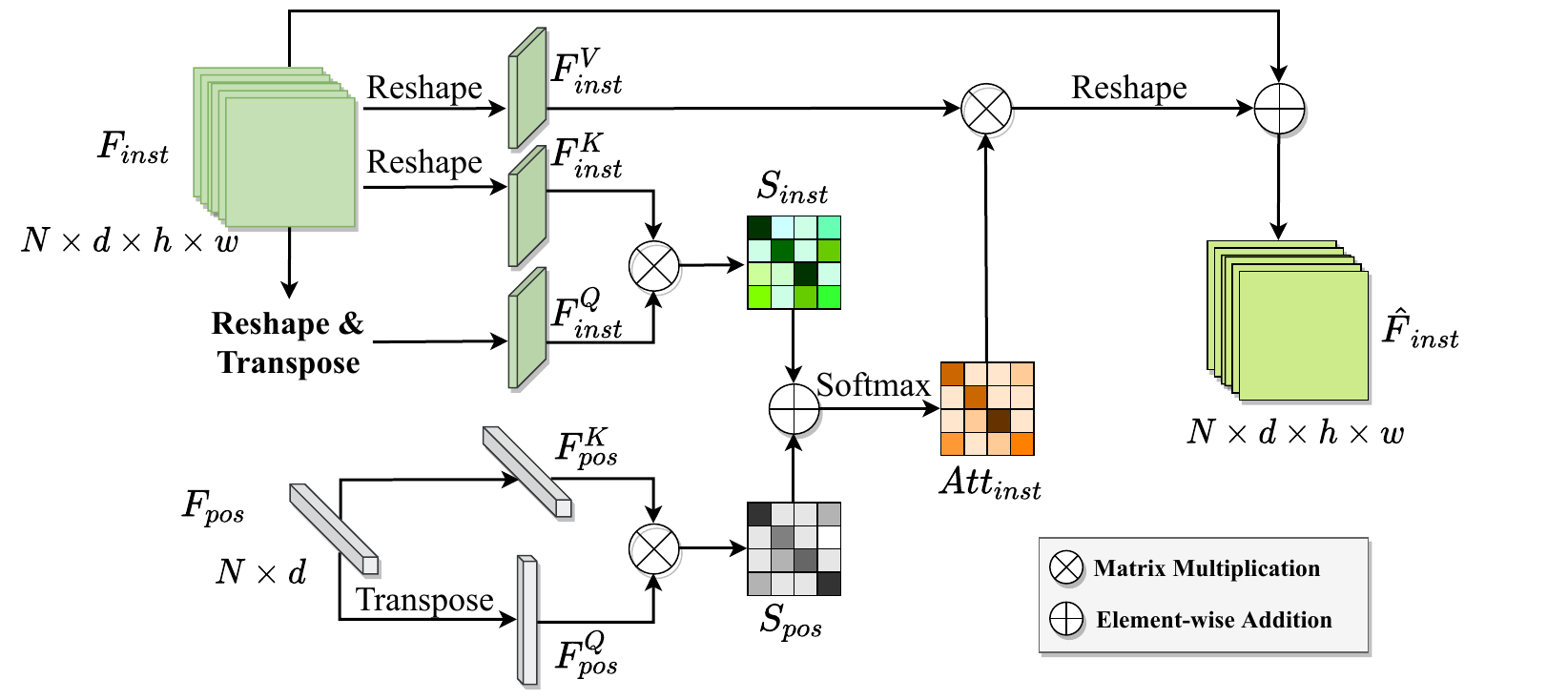}
    \caption{Structure of the proposed cross-instance interaction modeling module.}
    \label{fig:IRM}
\end{figure}

 \textbf{Input:} As shown in Figure \ref{fig:IRM}, the input to CIM includes the instance representation $F_{inst}$ and positional embedding $F_{pos}$ containing instances' position information. We incorporate $F_{pos}$ into CIM to enhance each person's positional information. Following \cite{CID}, we represent instances as a series of center maps and use the coordinate of the maxima as positional information.

 \textbf{Cross-instance correlation modeling:} To facilitate the modeling of apparent and positional correlations, we perform reshaping operation on $F_{inst}$ and $F_{pos}$ to generate three groups of instance representations ($F_{inst}^V \in \mathbb{R}^{N \times (d \times h \times w)}$, $F_{inst}^K\in\mathbb{R}^{N \times (d \times h \times w)}$, and $F_{inst}^Q \in \mathbb{R}^{ (d \times h \times w) \times N}$) and two groups of positional embeddings ($F_{pos}^K \in \mathbb{R}^{N \times d}$ and $F_{pos}^Q \in \mathbb{R}^{d \times N}$).

 The dot product models the similarity between two feature vectors, representing the degree of correlation between the vectors to some extent \cite{RPSTN}. We compute the dot product between feature vectors in $F_{inst}^K$ and $F_{inst}^Q$, $F_{pos}^K$ and $F_{pos}^Q$ to obtain apparent correlation $S_{inst} \in \mathbb{R}^{N \times N}$ and positional correlation $S_{pos}\in \mathbb{R}^{N \times N}$, respectively. Element-wise addition and softmax are applied to aggregate $S_{inst}$ and $S_{pos}$ resulting in the correlational attention map $Att_{inst} \in \mathbb{R}^{N \times N}$.
\begin{equation}
\begin{split}
    Att_{inst}^{ij} & = \frac{{\exp \left( {S_{inst}^{ij} + S_{pos}^{ij}} \right)}}{{\sum\limits_{j = 1}^N {\exp \left( {S_{inst}^{ij} + S_{pos}^{ij}} \right)} }} \\
    & = \frac{{\exp \left( {F_{inst}^{{K_i}} \cdot F_{inst}^{{Q_j}} + F_{pos}^{{K_i}} \cdot F_{pos}^{{Q_j}}} \right)}}{{\sum\limits_{j = 1}^N {\exp \left( {F_{inst}^{{K_i}} \cdot F_{inst}^{{Q_j}} + F_{pos}^{{K_i}} \cdot F_{pos}^{{Q_j}}} \right)} }}
\end{split}
\label{eq:correlation_att}
\end{equation}
 Each element $Att_{inst}^{ij}$ in $Att_{inst}$ measures the degree of interaction between any two instances. This allows the model to locate the current person's joints based on information from other people. The matrix multiplication between $Att_{inst}$ and $F_{inst}^V$ is performed to generate relation-based features $F_{inst}^R$.

 \textbf{Enhancement of instance-aware representations:} Because information about instance in $F_{inst}^R$ becomes confused after the dot product, we introduce the original instance representation $F_{inst}$ to enhance the discriminability of individual information.
 \begin{equation}
 \begin{split}
    \hat{F}_{inst} & = F_{inst}^R + {F_{inst}} \\
    & = {\mathop{\rm Re}\nolimits} {\rm{shape}}\left( {F_{inst}^V \cdot At{t_{inst}}} \right) + {F_{inst}}
 \end{split} 
 \label{eq:inst_residual}
 \end{equation}

\subsubsection{Cross-joint Interaction Modeling Module}
 \begin{figure}[htbp]
    \centering
    \includegraphics[scale=0.3]{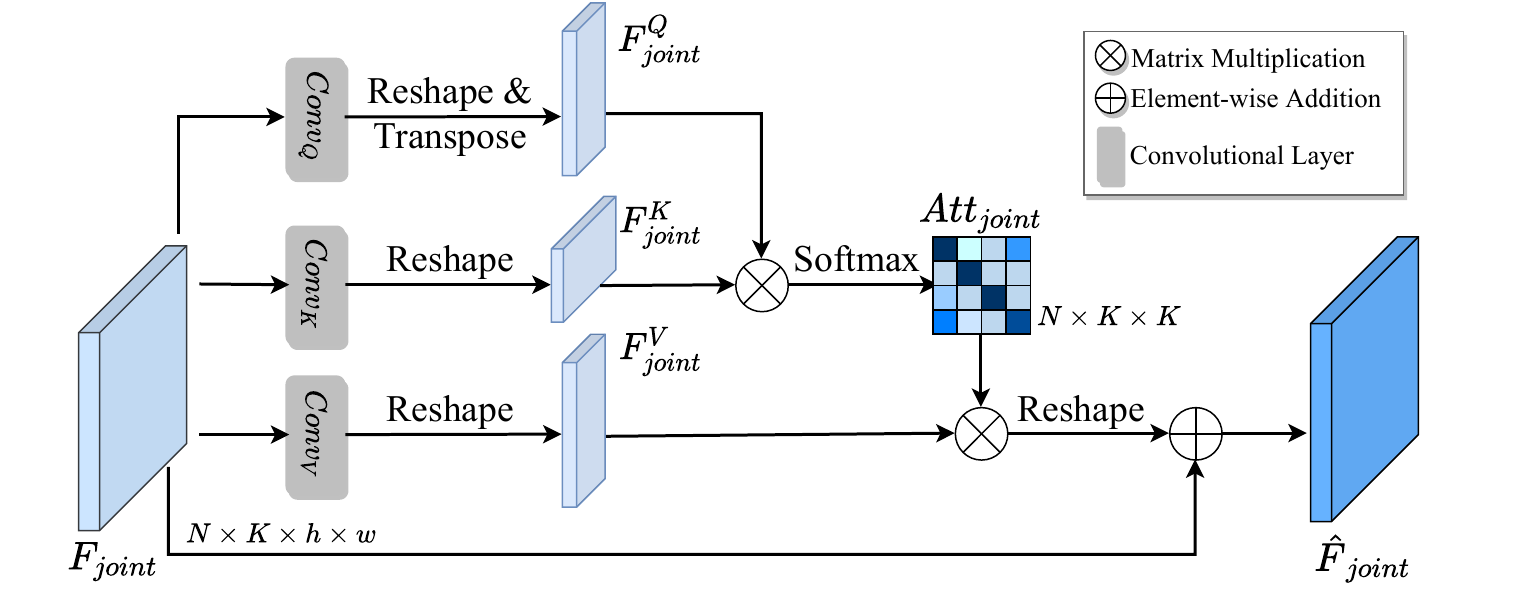}
    \caption{Structure of the proposed cross-joint interaction modeling module.}
    \label{fig:JRM}
 \end{figure}

 \textbf{Cross-joint correlation modeling:} As shown in Figure \ref{fig:JRM}, CJM takes joint features $F_{joint} \in \mathbb{R}^{N \times K \times h \times w}$ as input. Subsequently, three $1 \times 1$ convolutions, serving as joint feature extractors, are applied to $F_{joint}$ to generate three groups of discriminative joint features further. For the convenience of modeling the cross-joint correlation, reshaping and transpose operations are used to transform the joint feature matrices into feature vectors, \emph{i.e.}, $F_{joint}^Q \in \mathbb{R}^{N \times K \times (h \times w)}$, $F_{joint}^K \in \mathbb{R}^{N \times (h \times w) \times K}$, and $F_{joint}^V \in \mathbb{R}^{N \times K \times (h \times w)}$ as follows.
 \begin{equation}
    \begin{array}{l}
    F_{joint}^Q =  {{\rm{Reshape}}\left( {Con{v_Q}\left( {{F_{joint}}} \right)} \right)}   \\
    F_{joint}^K = {\rm{Trans}}\left( {\rm{Reshape}}\left( {Con{v_K}\left( {{F_{joint}}} \right)} \right) \right) \\
    F_{joint}^V = {\rm{Reshape}}\left( {Con{v_V}\left( {{F_{joint}}} \right)} \right)
    \end{array}
 \end{equation}
 Each row in $F_{joint}^Q$ and $F_{joint}^V$ or each row in $F_{joint}^K$ denotes a whole joint.

 Similar to cross-instance correlation modeling, dot product and softmax are used to calculate the spatial affinity between joints, generating the correlational attention map $Att_{joint} \in \mathbb{R}^{N \times K \times K}$.
 \begin{equation}
   Att_{joint}^{k,i} = \frac{{\exp \left( {F_{joint}^{Q_{(k,:)}} \cdot F_{joint}^{K_{(:,i)}}} \right)}}{{\sum\limits_{i = 1}^K {\exp \left( {F_{joint}^{Q_{(k,:)}} \cdot F_{joint}^{K_{(:,i)}}} \right)} }}
   \label{eq:joint_att}
 \end{equation}
 Each element $Att_{joint}^{k,i}$ in $Att_{joint}$ represents the degree of interaction between joint $k$ and $i$. This design allows the proposed DHRNet to locate the current joint based on information from other joints.

 \textbf{Enhancement of joint-aware representations:} Finally, we conduct a matrix multiplication between $Att_{joint}$ and $F_{joint}^V$ to consolidate information about joints, generating relation-based joint features. To enhance the discriminability of the individual joint information, we also incorporate the original joint feature $F_{joint}$ into the aggregated features, yielding the output feature $\hat{F}_{joint}$.
 \begin{equation}
    \hat{F}_{joint} = {\rm{Reshape}}\left(Att_{joint} \cdot F_{joint}^V \right) + F_{joint}
 \end{equation}

\subsubsection{Adaptive Feature Fusion Module} 
\label{sec:ADFM}
 The adaptive feature fusion module (ADFM), implemented based on channel-wise attention \cite{SENet}, is proposed to activate task-relevant features for pose estimation. Taking the IJR branch as an example, we first utilize global average pooling (GAP) to compress the spatial information of $\hat{F}_{inst}$ and $F_{joint}$ into a channel indicator. Then, a linear projection followed by a sigmoid non-linear function is employed to generate the channel-wise attention dedicated to activating significant instances or joint features. Finally, a  $1 \times 1$ convolution modulates and fuses the activated features. The entire process is formulated as follows.
 \begin{equation}
    \begin{array}{l}
    \tilde{F}_{joint} = {\rm Conv_d} \left( Att_c \odot \left[ \hat{F}_{inst}, F_{joint} \right] \right) \\
    Att_c = \sigma\left({\rm MLP}\left({\rm GAP} \left(\left[ \hat{F}_{inst}, F_{joint} \right] \right)\right)\right)
    \end{array}
    \label{eq:se_att}
 \end{equation}
 where $Att_c$ represents the channel-wise attention. $[\cdot,\cdot]$ and $\odot$ denote the concatenate operation and channel-wise multiplication, respectively. $\sigma(\cdot)$, MLP($\cdot$), and GAP($\cdot$) refer to the sigmoid function, linear projection, and global average pooling.

\subsection{Pose Decoder \& Training Loss}

 \textbf{Pose decoder:} To obtain robust pose features, we introduce the channel and spatial attention \cite{CBAM} into the pose decoder to adaptively highlight task-relevant features. Then, two convolutions are used to generate joint heatmaps, as follows.
 \begin{equation}
    \begin{array}{l}
      \tilde{F} = \mathcal{SA}\left( \mathcal{CA}\left( \left[ F^{I \rightarrow J}, F^{J \rightarrow I} \right] \right) \right) \\
      \mathcal{H} = {\rm Conv} \left({\rm Conv}\left( \tilde{F}\right)\right)
    \end{array}
    \label{eq:decoder}
 \end{equation}
 where $\tilde{F}$ denotes the pose features enhanced by integrating interactive information from different objects. $\mathcal{SA(\cdot)}$ and $\mathcal{CA(\cdot)}$ refer to the spatial-wise and channel-wise attention mechanism in CBAM \cite{CBAM}. $F^{I \rightarrow J}$ and $F^{J \rightarrow I}$ are the relation-based features extracted by IJR and JIR branches, respectively.

 \textbf{Training loss:} Following \cite{CID}, the loss function includes an instance segmentation loss used to extract robust instance representations and a joint heatmap loss applied to force the model to learn task-relevant information.
 \begin{equation}
    \mathcal{L}_{total} = \mathcal{L}_{inst} + \alpha \cdot \mathcal{L}_{joint}
 \end{equation}
 where $\alpha$ is a balanced factor to adjust the importance of different loss terms. $\mathcal{L}_{inst}$ is the Focal loss between the center maps of instances output by the model and the ground truth \cite{CID}. $\mathcal{L}_{joint}$ is the mean square error between the joint heatmaps output by the model and the ground truth.

\section{Experiments}
 
 This section commences with an overview of the dataset employed, including details on evaluation metrics and implementation specifics. Subsequently, we conduct a comparative analysis of the proposed DHRNet against state-of-the-art methods. Following this, we present an extensive ablation experiment delving into the various components of our proposed method. Finally, we showcase visual results and offer a qualitative analysis of the effectiveness of our approach.
 
\subsection{Datasets}
We evaluate the proposed method on the COCO \cite{COCO}, CrowdPose\cite{CrowdPose}, and OCHuman\cite{OCHuman} datasets. 

\textbf{COCO keypoint dataset} \cite{COCO} contains 64K images with 270K instances annotated with 17 body joints. Following \cite{HRNet,CID}, we train the proposed method on the COCO train2017 set, including 57K images and 150K persons. Moreover, we evaluate our approach on the COCO val2017 and test-dev2017 sets, respectively.

\textbf{CrowdPose dataset} \cite{CrowdPose} contains 80K human poses with 14 labeled body joints. Following \cite{CID}, we use the travel set, including 12K images, to train our model and use the test set, including 8K images, to evaluate the model.

\textbf{OCHuman dataset} \cite{OCHuman} contains 8110 annotated instances within 4731 images. It provides 17 labeled body joints for each instance. Following \cite{CID}, we adopt two different experimental settings to validate the proposed method.

\subsection{Evaluation metric} 
Following \cite{CID,HigherHRNet}, we adopt the average precision (AP)  as the evaluation metric and report the AP, ${\rm AP}^{50}$, and ${\rm AP}^{75}$ to show the joint positioning accuracy under different thresholds. Moreover, on COCO, we report${\rm AP}^{M}$ and ${\rm AP}^{L}$, which refers to the performance for multi-scale objects. On CrowdPose, we report  ${\rm AP}^{E}$, ${\rm AP}^{M}$, and ${\rm AP}^{H}$ to validate the performance under different crowd indexes.

\subsection{Implementation Details}

\textbf{Structure:} We use the HRNet-W32 \cite{HRNet} as feature encoder, and GFD \cite{CID} as instance and joint decoders to decouple instances' and joints' features. 

\textbf{Training Details:} The input size and data augmentation are consistent with \cite{CID}. The batch size is also set to 20 for OCHuman and 32 for CrowdPose and COCO datasets. We train the model 60 epochs on the COCO dataset with the learning rate divided by ten at $20^{th}$ and $40^{th}$ epoch. We train the model 100 epochs on the CrowdPose dataset with the learning rate decreased by ten at $40^{th}$ and $70^{th}$ epoch. For the OCHuman dataset, we train the model 80 epochs with the learning rate decreased by ten at $40^{th}$ and $60^{th}$ epoch.

Consistent with \cite{CID}, we resize the input image to $512 \times 512$; use random rotation (-30,30), scale ([0.75, 1.5]), translation ([-40, 40]), and flipping as data augmentation strategies; set the initial learning rate to 0.001; choose Adam \cite{Adam} as the optimizer to train the model. In addition, the batch size is set to 20 for OCHuman and 32 for CrowdPose and COCO datasets since the data volume of the OCHuman is smaller than that of the CrowdPose and COCO. We train the model 60 epochs on the COCO dataset with the learning rate divided by ten at $20^{th}$ and $40^{th}$ epoch. We train the model 100 epochs on the CrowdPose dataset with the learning rate decreased by ten at $40^{th}$ and $70^{th}$ epoch. For the OCHuman dataset, we train the model 80 epochs with the learning rate decreased by ten at $40^{th}$ and $60^{th}$ epoch.

\subsection{Comparison with State-of-the-art Methods}

\subsubsection{Results on the COCO dataset}
Experimental results are reported in Table \ref{tab:coco}. We record parameters, GFLOPs, average inferencing speed, and performance. Besides, each group's best results are marked with bold text, and second-best results are underlined. 

Table \ref{tab:coco} illustrates that DHRNet surpasses bottom-up and single-stage methods while introducing only a minimal parameter increase. Notably, DHRNet achieves a 0.3\% improvement over the previous state-of-the-art CID \cite{CID}, underscoring the value of spatial correlation information in multi-person scenarios for joint localization. Despite a 5.57 GFLOPs increase relative to CID \cite{CID}, DHRNet exhibits a mere 0.2-second average inference speed decrement. However, it is worth noting that DHRNet demonstrates inferior performance compared to top-down methods like SBL and HRNet. This discrepancy may be attributed to the heightened sensitivity of single-stage methods to multi-scale targets, a problem mitigated by the first-stage detector in top-down approaches.

\begin{table*}[tbp]
    \footnotesize
    \centering
    \setlength\tabcolsep{5.5pt}
    \caption{Comparisons with the state-of-the-art methods on the COCO test-dev set. The \textbf{best results} are in \textbf{bold}, and the \underline{second best results} are \underline{underlined} in each group. $\ddag$ denotes the approach reproduced on our experimental platform.}
\begin{tabular}{l|c|c|c|c|c|ccccc|c}
    \hline
    Methods & Backbone & Input size & Params (M) & GFLOPs & Speed (s) & ${\rm AP}$ & ${\rm AP^{50}}$ &  ${\rm AP^{75}}$ & ${\rm AP^{M}}$ &  ${\rm AP^{L}}$ & AR \\
    \hline
    \multicolumn{10}{c}{Top-down methods} \\
    \hline
    
    Mask R-CNN \cite{MaskR-CNN}     & ResNet-50 & 800 & - & - & - & 63.1 & 87.3 & 68.7 & 57.8 & 71.4 & - \\
    ${\rm SBL^\dagger}$\cite{SBL}                  & ResNet-50 & $384\times288$ & 34.0 & - & - &\underline{70.4} & \underline{88.6} & \underline{78.3} & \underline{67.1} & \underline{77.2} & - \\
    ${\rm HRNet^\dagger}$\cite{HRNet}   & HRNet-W32 & $384 \times 288$ & 28.5 & - & - & \textbf{74.4} & \textbf{90.5} & \textbf{81.9} & \textbf{70.8} & \textbf{81.0} & - \\
    \hline
    \multicolumn{10}{c}{Bottom-up methods} \\
    \hline
   
    OpenPose\cite{Openpose}         & VGG-19 & - & - & - & 0.1 & 61.8 & 84.9 & 67.5 & 57.1 & 68.2 & 66.5 \\
    AE\cite{AE}                     & Hourglass & 512 & - & - & 0.12 & 62.8 & 84.6 & 69.2 & 57.5 &  70.6 & - \\
    PersonLab\cite{PersonLab}       & ResNet-152 & 1401 & - & - & 0.2 & 66.5 & 88.0 & 72.6 & \textbf{62.4} & 72.3 & 71.0 \\
    HrHRNet\cite{HigherHRNet}       & HRNet-W32 & 512 & 28.6 & - & 0.21 & 66.4 & 87.5 & 72.8 & 61.2 & 74.2 & - \\
    CA \cite{CenterAttention} & HrHRNet-W32 & 512 & - & - & - & 67.6 & 88.7 & 73.6 & 61.9 & \underline{75.6} & - \\
    SWAHR\cite{SWAHR}               & HrHRNet-W32 & 512 & 28.6 & - & 0.21 & \underline{67.9} & \textbf{88.9} & \underline{74.5} & \underline{62.4} & 75.5 & - \\
    DecenterNet\cite{decenternet}   & HRNet-W32 & 512 & 29.7 & 48.5 & - & \textbf{69.0} & \underline{87.9} & \textbf{75.8} & \textbf{63.3} & \textbf{77.3} & - \\
    
    \hline
    \multicolumn{10}{c}{Single-stage methods} \\
    \hline
    CenterNet\cite{CenterNet}       & Hourglass & 512 & 63.0 & - & - & 86.8 & 69.6 & 58.9 & 70.4 & - \\
    SPM\cite{SPM}                   & - & - & - & - & - & 66.9 & 88.5 & 72.9 & 62.6 & 73.1 & - \\
    PointSet\cite{PointPose}        & HRNet-W48 & 800 & 66.3 & - & 0.21 & 87.7 & 73.4 & 64.9 & 70.0 & - \\
    FCPose\cite{FCPose}             & ResNet-101 & 800 & - & - & - & 65.6 & 87.9 & 72.6 & 62.1 & 72.3 & - \\
    InsPose\cite{InsPose}           & ResNet-101 & - & - & - & - & 66.3 & 89.2 & 73.0 & 61.2 & 73.9 & - \\
    DEKR \cite{DEKR}                & HRNet-W32 & 512 & 29.6 & - & 0.2 & 67.3 & 87.9 & 74.1 & 61.5 & 76.1 & 72.4 \\
    CID$^\ddag$ \cite{CID}          & HRNet-W32 & 512 & 29.36 & 75.88 & 0.12 & \underline{68.7} & \textbf{89.8} & \underline{75.7} & \underline{63.0} & \textbf{77.5} & \underline{74.3} \\
    \textbf{DHRNet (Ours)}         & HRNet-W32 & 512 & 29.37 & 81.45 & 0.33 & \textbf{69.0}  & \textbf{89.8} & \textbf{76.4} & \textbf{63.5}  & \textbf{77.5} & \textbf{74.7}  \\
    \hline
    \textbf{Performance Gain}      & - & - & \textcolor{blue}{-0.01} & \textcolor{blue}{-5.57} & \textcolor{blue}{-0.21} & \textcolor{red}{+0.3} & - & \textcolor{red}{+0.7} & \textcolor{red}{+0.5} & - & \textcolor{red}{+0.4} \\
   \hline
   \end{tabular}
   \label{tab:coco}
 \end{table*}

\subsubsection{Results on the CrowdPose dataset}
To assess the effectiveness of the proposed DHRNet, we conduct a comparative analysis with existing methods using the CrowdPose dataset, as summarized in Table \ref{tab:crowdpose}. Overall, DHRNet achieves a remarkable 71.5\% Average Precision (AP), outperforming other methods comprehensively. Furthermore, across various levels of crowded scenarios, DHRNet consistently outperforms current State-of-the-Art (SOTA) results. These findings underscore the efficacy of interaction modeling in enhancing the model's performance. By leveraging correlation information at different levels, DHRNet effectively utilizes auxiliary information to locate human joint points accurately.
\begin{table}[tbp]
    \centering
    \setlength\tabcolsep{3.1pt}
    \caption{Comparisons with state-of-the-art methods on the CrowdPose test-dev. The \textbf{best results} are in \textbf{bold} and the \underline{second best results} are \underline{underlined}. $\ddag$ denotes the approach reproduced on our experimental platform.}
    \begin{tabular}{l|ccc|ccc}
    \hline
    Methods & ${\rm AP}$ &  ${\rm AP^{50}}$ &  ${\rm AP^{75}}$ & ${\rm AP^{E}}$ & ${\rm AP^{M}}$ &  ${\rm AP^{H}}$  \\
    \hline
    \multicolumn{6}{c}{Top-down methods} \\
    \hline
    
    Mask R-CNN \cite{MaskR-CNN}     & 57.2 & 83.5 & 60.3 &  69.4 & 57.9 & 45.8 \\
    JC-SPPE \cite{CrowdPose}        & 66.0 & 84.2 & 71.5 & 75.5 & 66.3 & 57.4 \\
    \hline
    \multicolumn{6}{c}{Bottom-up methods} \\
    \hline
   
    OpenPose \cite{Openpose}         &   -  & -    & -    & 62.7 & 48.7 &  32.3 \\
    HrHRNet \cite{HigherHRNet}  & 65.9 & 86.4 & 70.6 & 73.3 & 66.5 & 57.9 \\
    CA \cite{CenterAttention} & 67.6 & 87.7 & 72.7 &  68.1 & 75.8 & 58.9 \\
    DecenterNet \cite{decenternet}& 69.3 & - & - & 76.8 & 70.0 & 60.8 \\
    \hline
    \multicolumn{6}{c}{Single-stage methods} \\
    \hline
    DEKR \cite{DEKR}      & 65.7 & 85.7 & 70.4 & 73.0 & 66.4 & 57.5 \\
    PINet \cite{PINet}    & 68.9 & 88.7 & 74.7 & 75.4 & 69.6 & 61.5 \\
    SPL \cite{SPL}        & 69.0  & - & - & 76.2 & 70.0 & 60.4 \\
    CID$^\ddag$ \cite{CID}        & \underline{71.2} & \underline{89.8} & \underline{76.7} & \underline{77.4} & \underline{71.9} & \underline{63.8} \\
    \textbf{DHRNet (Ours)}         & \textbf{71.5}  & \textbf{90.3} & \textbf{77.1} & \textbf{77.9} & \textbf{72.2} & \textbf{64.0}  \\
    \hline

    \textbf{Performance Gain}      & \textcolor{red}{+0.3} & \textcolor{red}{+0.5} & \textcolor{red}{+0.4} & \textcolor{red}{+0.5} & \textcolor{red}{+0.3} & \textcolor{red}{+0.2} \\
   \hline
   \end{tabular}
   \label{tab:crowdpose}
 \end{table}

\subsubsection{Results on the OCHuman dataset}
In order to validate DHRNet's effectiveness, we evaluate it on the OCHuman dataset, which consists of severely occluded scenes. Experimental results are listed in Table \ref{tab:ochuman}. Following \cite{CID}, we also adopt two experimental settings: training the model on the OCHuman val and the COCO datasets, respectively, and evaluating the model on the OCHuman testing dataset. 

For the first setting, DHRNet outperforms CID by 1.0\% AP. For the second setting, our DHRNet also outperforms CID by 0.3\% AP on the OCHuman val set and 0.8\% AP on the testing set. It is evident in Table \ref{tab:ochuman} that, equipped with the DIM, the model's performance is significantly improved.

\begin{table}[htbp]
    \centering
    \caption{Comparisons with the state-of-the-art methods on the OCHuman dataset. The \textbf{best results} are in \textbf{bold} and the \underline{second best results} are \underline{underlined}. $\ddag$ denotes the approach reproduced on our experimental platform.}
    \begin{tabular}{l|c|c}
    \hline
    Methods & OCHuman val & COCO train \\
    \cline{2-3}
    \hline
    \multicolumn{3}{c}{Top-down methods} \\
    \hline
    Mask R-CNN \cite{MaskR-CNN}  & 20.2 & - \\
    JC-SPPE \cite{CrowdPose}     & 27.6 & - \\
    OPEC-Net \cite{OPEC-Net}     & 29.1 & - \\
    RMPE \cite{RMPE}             & -    & 30.7 \\
    SBL \cite{SBL}               & -    &  33.3 \\
    \hline
    \multicolumn{3}{c}{Bottom-up methods} \\
    \hline
    AE \cite{AE}                 & -    & 29.5 \\
    HGG \cite{HGG}               & -    & 34.8 \\
    HrHRNet \cite{HigherHRNet}   & 27.7 & 39.4 \\
    \hline
    \multicolumn{3}{c}{Single-stage methods} \\
    \hline
    SPM \cite{SPM}               & 45.6 & - \\
    DEKR \cite{DEKR}             & 52.2 & 36.5 \\
    CID$^\ddag$\cite{CID}        & \underline{57.5} & \underline{44.0} \\
    \textbf{DHRNet (Ours)}      & \textbf{58.5} (\textcolor{red}{+1.0}) & \textbf{44.8} (\textcolor{red}{+0.8}) \\
    \hline
    \end{tabular}
    \label{tab:ochuman}
\end{table}

\subsection{Ablation Studies}

In this section, we first evaluate the effectiveness of the dual-path relation modeling. Next, we validate the different combinations of relational blocks and the influence of the adaptive feature fusion module. We conduct a series of ablation studies on the OCHuman dataset\cite{OCHuman}.

\begin{table}[htbp]
    \footnotesize
    \setlength\tabcolsep{2pt}
    \caption{Ablation studies about various designs of DIM on the OCHuman dataset. Subscripts "IJ" and "JI" represent instance-joint and joint-instance relational branches, respectively.}
    \centering
    \begin{tabular}{c|cccc|cccc}
        \hline
         Method & CIM$_{IJ}$ & CJM$_{IJ}$ & CJM$_{JI}$ & CIM$_{JI}$ & AP & AP$^{50}$ & AP$^{75}$ & AR  \\
        \hline
         Baseline & & & & & 56.3 & 74.6 & 61.4 & 74.7 \\
         (a)    & \checkmark & \checkmark &            &            & 58.0 & 75.8 & 63.5 & 75.7  \\
         (b)    &            &            & \checkmark & \checkmark & 57.9 & 75.6 & 63.4 & 75.8  \\
         \hline
         (c)    & \checkmark &            &  & \checkmark  & 56.7 & 74.9 & 61.7 & 74.6  \\
         (d)    &            & \checkmark & \checkmark  &  & 57.2 & 75.6 & 62.6 & 75.0  \\
        \hline
        \textbf{DHRNet}    & \checkmark & \checkmark & \checkmark & \checkmark & \textbf{58.6} & \textbf{76.4} & \textbf{64.4} & \textbf{76.5}  \\
        \hline
    \end{tabular}
    \label{tab:DHRM_ablation}
\end{table} 

\subsubsection{Studies on Different Branches of DHRM} 
We explore the influence of single-branch relation modeling with the fixed order. As shown in TABLE \ref{tab:DHRM_ablation}, $Baseline$ denotes the DHRNet without DIM. $(a)$ is the method that DHRNet is only equipped with the instance-joint relational branch (IJR), while $(b)$ refers to the DHRNet equipped with the joint-instance relational branch (JIR). Compared with $Baseline$, both relation modeling branches bring improvement for the model, indicating that correlational information is helpful for pose estimation. Moreover, the JIR branch achieves comparable performance with the IJR branch (58.0\% AP vs 57.9\% AP), indicating that both complementary relational modeling orders play an essential role in multi-person pose estimation. The model's performance is further improved when both relation modeling exists, \emph{i.e.}, DHRNet.

\subsubsection{Studies on Single-level Interaction Modeling} 

We combine the same interaction modeling modules in two branches to validate the effect of the single-level correlation. Experimental results are recorded as $(c)$ DHRNet with two CIMs and $(d)$ DHRNet with two CJMs. After introducing the single-level interaction modeling, the performance of $(c)$ and $(d)$ outperform $Baseline$. However, they are inferior to multi-level interaction modeling approaches since single-level relation modeling only extracts onefold correlations.

\subsubsection{Studies on Adaptive Feature Fusion Module}
We remove the ADFM from DIMM and pose decoder, respectively, to study the effectiveness of ADFM, as shown in TABLE \ref{tab:ADFM}. $Baseline$ denotes the method without ADFMs in both DIM and pose decoder. Method $(a)$ represents the method only with ADFM in DIM, and $(b)$ refers to the approach only with ADFM in the pose decoder.

Method $(a)$ outperforms $Baseline$ by 0.3\% AP since ADFM in DHRM can help the model pay attention to the important features. Similarly, ADFM$_D$ guides the model in activating the salient and significant features from the output of two interaction modeling branches.
\begin{table}[htbp]
    \footnotesize
    \centering
    \caption{Ablation studies about the adaptive feature fusion modules in DIM and pose decoder.}
    \begin{tabular}{c|cc|cccc}
    \hline
        Method      & ADFM       & ADFM$_{D}$ & AP & AP$^{50}$ & AP$^{75}$ & AR \\
    \hline
        Baseline    &            &                  & 57.6 & 75.7 & 62.8 & 75.3 \\
        (a)         & \checkmark &                  & 57.9 & 75.9 & 62.8 & 75.7 \\
        (b)         &            & \checkmark       & 58.1 & 76.2 & 63.9 & 76.2 \\
        \textbf{Ours} & \checkmark & \checkmark       & \textbf{58.6} & \textbf{76.4} & \textbf{64.4} & \textbf{76.5} \\
    \hline
    \end{tabular}
    \label{tab:ADFM}
\end{table}

\subsection{Qualitative Analysis}

\subsubsection{Visualization of Instance-level Correlations} 
\begin{figure}[htbp]
    \centering
    \includegraphics[scale=0.21]{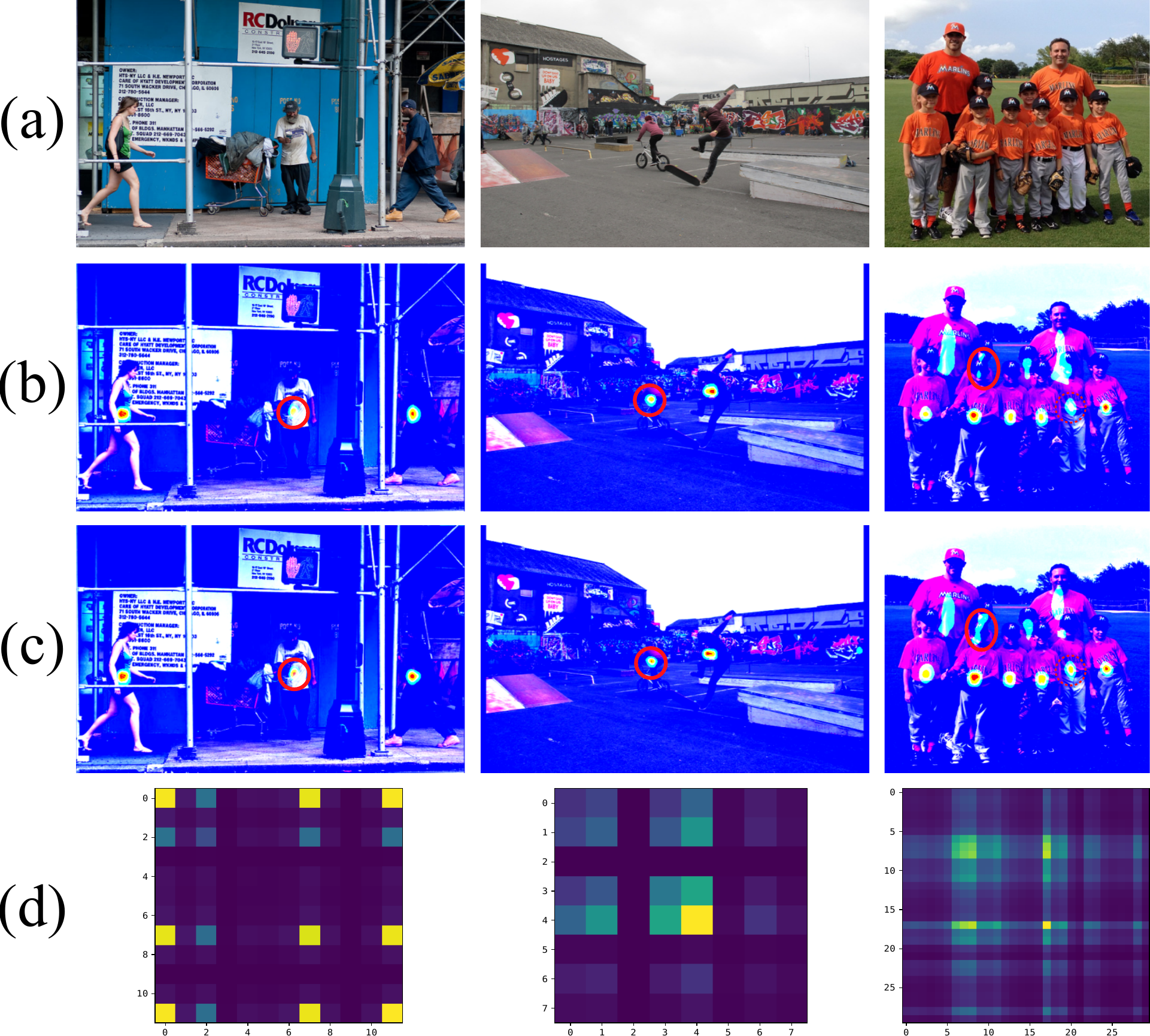}
    \caption{Visualization about center maps and correlations on the COCO val dataset. (a) represents the output of CID. (b) denotes the output of our DHRNet. (c) and (d) are cross-instance correlations in the IJR and JIR branches. In Figure (d), the x-axis and y-axis represent the index of each person's proposal.}
    \label{fig:multi-instance-correlation}
\end{figure}

Figure \ref{fig:multi-instance-correlation} illustrates the cross-instance correlations alongside instances' center maps. In Figure (d), the x-axis and y-axis represent the indices of each person's proposal. Typically, to enhance recall, the number of proposals exceeds the number of bounding boxes for people. The cross-instance correlation maps depicted in Figure (d) showcase correlations between individuals. For instance, in the middle correlation map of Figure (d), a person with index 4 exhibits correlations not only with themselves but also with the proposals of individuals indexed 1 and 3. These additional proposals (1st and 3rd) provide supplementary information aiding in the localization of the person indexed 4. Leveraging its richer auxiliary properties, DHRNet demonstrates superior performance compared to CID, as evidenced by features such as the center map highlighted in red.

\subsubsection{Visualization of Joint-level Correlations}

\begin{figure}[htbp]
    \centering
    \includegraphics[scale=0.125]{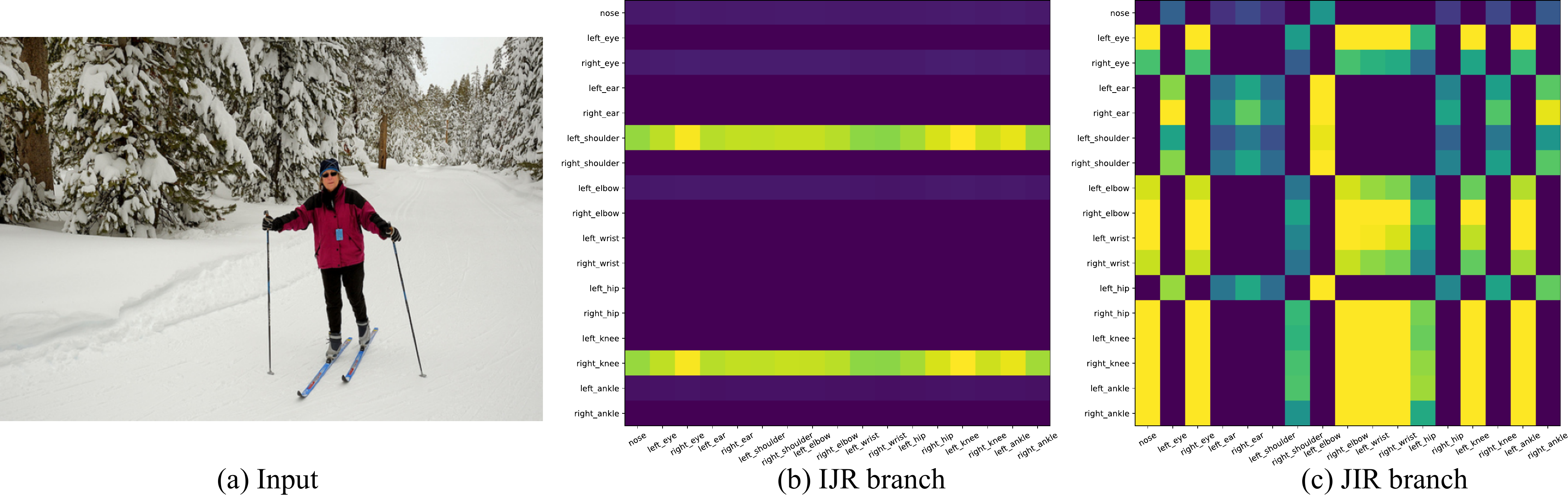}
    \caption{Visualization of correlations between joints. Figure (a) refers to the input image. Figures (b) and (c) represent cross-joint correlations in the IJR and JIR branches.}
    \label{fig:joint-correlation}
\end{figure}

Figure \ref{fig:joint-correlation} shows cross-joint correlations in the IJR and JIR branches. Each element in correlational maps represents the degree of correlation between any two joints. Since the input of CJM in the IJR branch incorporates instances' information, the joint-level correlation learned by IJR is coarse-grained. The input of CJM in the JIR branch contains distinguishing information about joints, so the joint-level correlation obtained by JIR is fine-grained. 

\subsubsection{Features of IJR and JIR Branches}
\begin{figure*}
    \centering
    \includegraphics[scale=0.45]{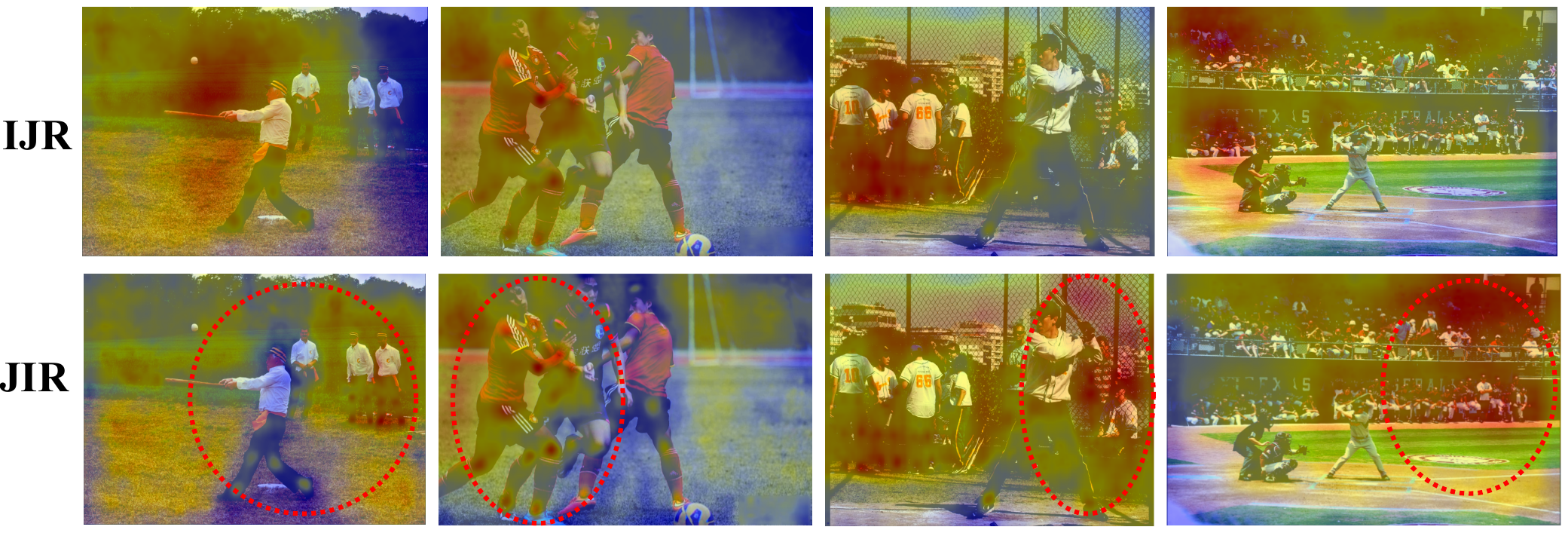}
    \caption{Output features of IJR and JIR branches. The intensity of orange shading corresponds to the level of attention, with darker orange indicating higher attention values. Conversely, blue regions signify lower attention.}
    \label{fig:attention}
\end{figure*}

As shown in Figure \ref{fig:attention}, we present a visual representation of the output from the IJR and JIR branches, allowing for an intuitive observation of their attentional patterns. The Visualization reveals a complementary relationship between the attention mechanisms of the IJR and JIR branches. While the IJR branch focuses on certain regions, the JIR branch attends to areas overlooked by the former. This complementary behavior suggests a collaborative effort between these two branches, enhancing the model's scene comprehension. Experimental results corroborate the significance of this complementary relation in the context of multi-person pose estimation.

\subsubsection{Visualization of Human Poses}

\begin{figure*}[htbp]
    \centering
    \includegraphics[scale=0.33]{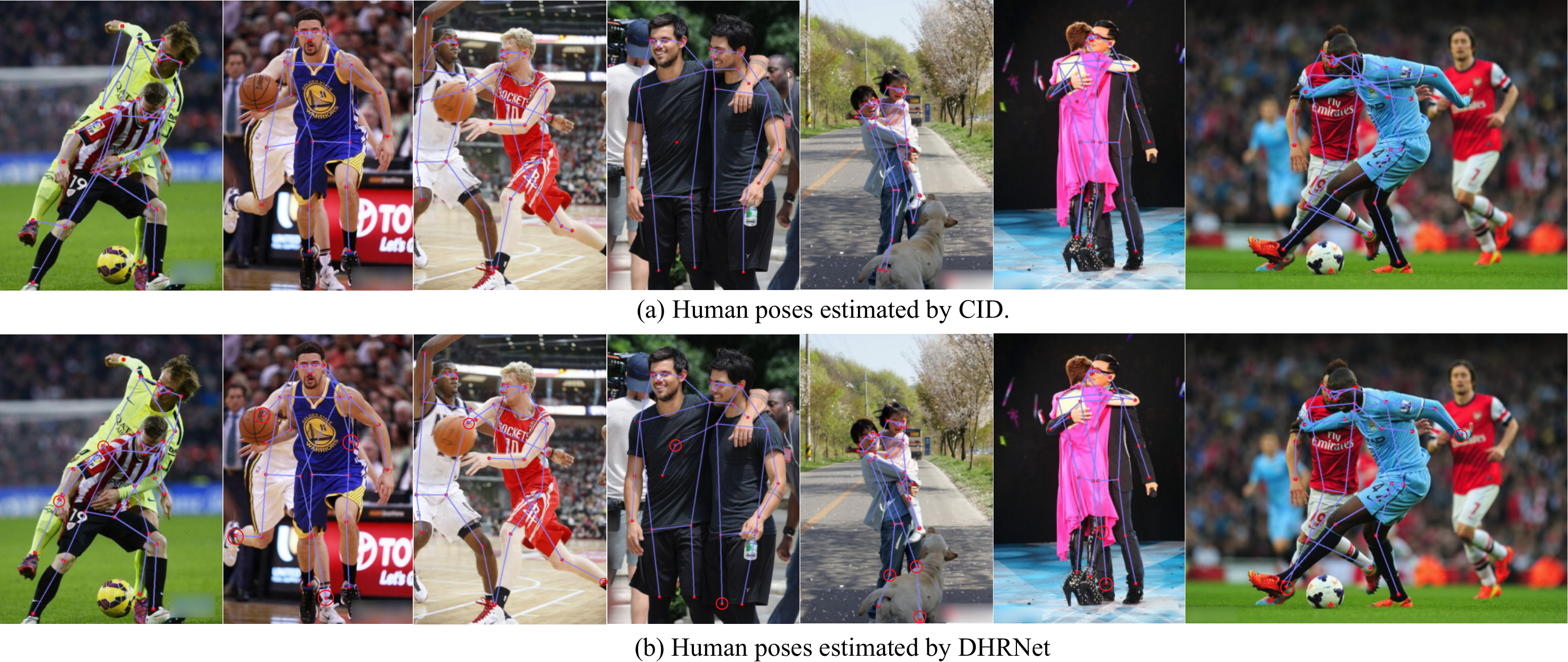}
 \caption{Qualitative results on the OCHuman dataset. (a) shows the qualitative results of CID. (b) shows the qualitative results of our DHRNet. We mark the joints in red circles where DHRNet outperforms CID.}
\label{fig:occlud_vis}
\end{figure*}

\textbf{Visualization on the OCHuman dataset.} We visualize the pose estimation results of DHRNet and CID on the OCHuman dataset to evaluate the model's performance on the occluded cases, as shown in Figure \ref{fig:occlud_vis}. For the location of occluded joints (joints in the red circles), DHRNet outperforms CID. With the guidance of dual-path interaction modeling, DHRNet can capture interactions between instances by cross-instance correlation and the pose structure information by cross-joint correlation, which allows DHRNet to locate the occluded joint based on information from other instances and joints.

\begin{figure*}[htbp]
    \centering
    \includegraphics[scale=0.7]{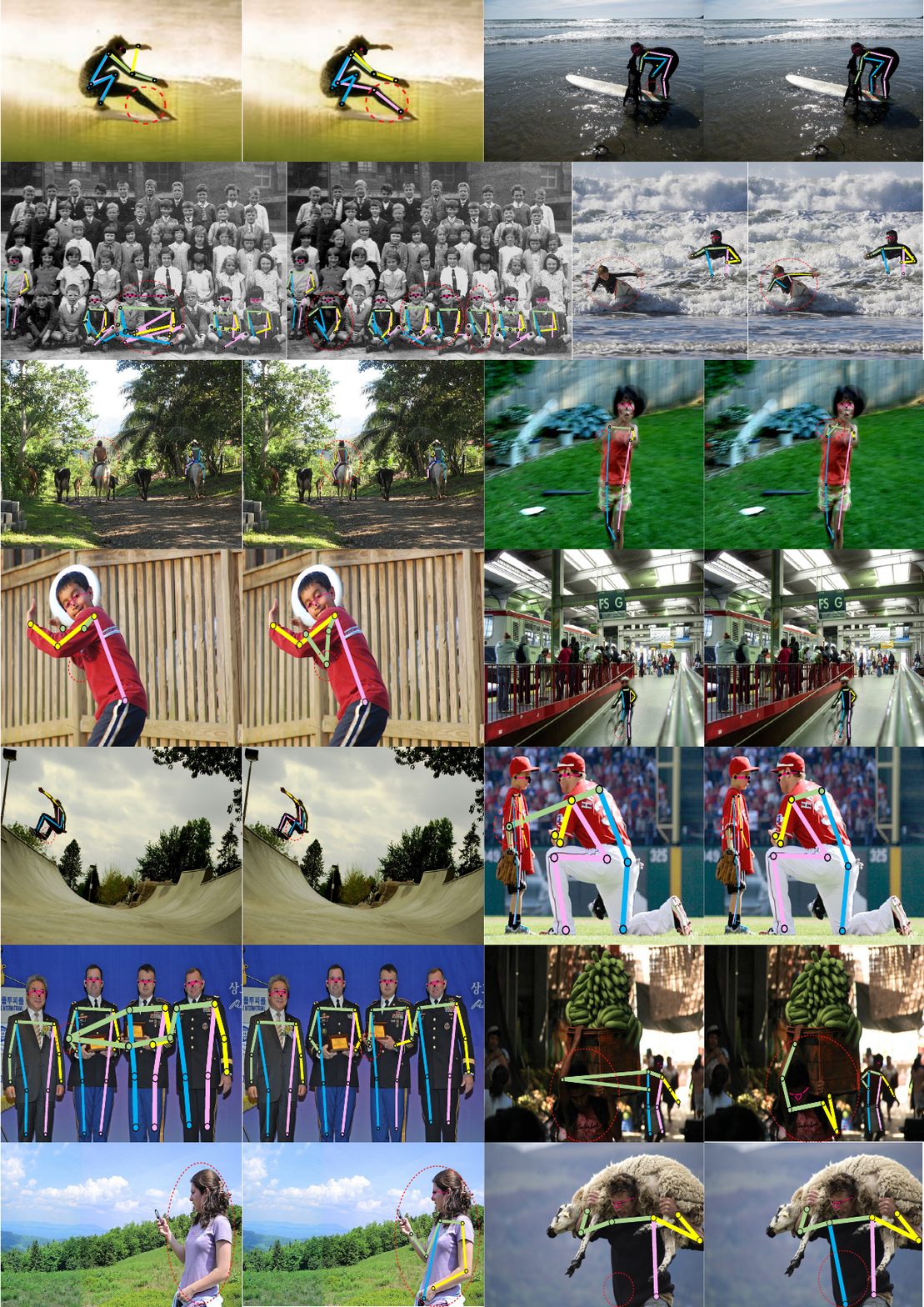}
    \caption{Visualization of human poses on the COCO dataset. From left to right, each example shows the results of CID and DHRNet. The incorrect regions are marked with red circles.}
    \label{fig:coco_pose}
\end{figure*}

 \textbf{Visualization on the COCO dataset.} Figure \ref{fig:coco_pose} shows the qualitative results on the COCO dataset \cite{COCO}. Paired visualizations present the output of CID \cite{CID} (the left figure) and our DHRNet (the right figure). For some complex poses, CID fails to detect keypoints, while DHRNet can accurately locate its joint points under the guidance of interactive information. Furthermore, the joints predicted by CID are prone to misalignment due to insufficient cross-instance and cross-joint interaction modeling. Experimental results show that the proposed DHRNet is robust against occlusion and complex poses, proving our method's effectiveness.

\section{Conclusion}

This work proposes a novel dual-path hierarchical relation modeling network (DHRNet) to capture interactions between instances or joints flexibly and autonomously. DHRNet includes a key component: a dual-path interaction modeling module (DIM). By integrating the information of other instances and joints related to the current person, DHRNet obtains richer auxiliary information for estimating the current person's pose. We hope this work will benefit the community by highlighting the importance of cross-instance and cross-joint interactions.

\section*{Acknowledgments}
 This work was supported partly by the National Natural Science Foundation of China (Grant No. 62173045, 62273054), partly by the Fundamental Research Funds for the Central Universities (Grant No. 2020XD-A04-3), and the Natural Science Foundation of Hainan Province (Grant No. 622RC675).
%


\bibliographystyle{model1-num-names}

\bibliography{cas-refs}



\end{document}